\documentclass[sn-mathphys-num]{sn-jnl}

\usepackage{graphicx}%
\usepackage{multirow}%
\usepackage{amsmath,amssymb,amsfonts}%
\usepackage{amsthm}%
\usepackage{mathrsfs}%
\usepackage[title]{appendix}%
\usepackage{xcolor}%
\usepackage{textcomp}%
\usepackage{manyfoot}%
\usepackage{booktabs}%
\usepackage{algorithm}%
\usepackage{algorithmicx}%
\usepackage{algpseudocode}%
\usepackage{listings}%
\usepackage{enumitem}

\theoremstyle{thmstyleone}%
\theoremstyle{thmstyletwo}%
\newtheorem{example}{Example}%

\theoremstyle{thmstylethree}%
\newtheorem{definition}{Definition}%

\def\mcl#1{\mathcal{#1}}

\def\hil{\mcl{H}}

\def\alg{\mcl{A}}

\def\bra#1{\langle#1\vert}
\def\ket#1{\vert#1\rangle}
\def\qbracket#1#2{\langle#1\vert#2\rangle}

\DeclareSymbolFont{EulerExtension}{U}{euex}{m}{n}
\DeclareMathSymbol{\euintop}{\mathop} {EulerExtension}{"52}
\DeclareMathSymbol{\euointop}{\mathop} {EulerExtension}{"48}

\usepackage{booktabs}
\usepackage[capitalize]{cleveref}

\begin{document}

\title[]{Quantum Circuit $C^*$-algebra Net}


\author*[1,2]{\fnm{Yuka} \sur{Hashimoto}}\email{yuka.hashimoto@ntt.com}
\equalcont{These authors contributed equally to this work.}

\author[2]{\fnm{Ryuichiro} \sur{Hataya}}\email{ryuichiro.hataya@riken.jp}
\equalcont{These authors contributed equally to this work.}

\affil*[1]{\orgdiv{NTT Network Service Systems Laboratories}, \orgname{NTT Corporation}, \orgaddress{\street{3-9-11, Midoricho}, \city{Musashino-shi}, \postcode{1808585}, \state{Tokyo}, \country{Japan}}}

\affil[2]{\orgdiv{Center for Advanced Intelligence Project (AIP)}, \orgname{RIKEN}, \orgaddress{\street{Nihonbashi 1-chome Mitsui Building, 15th floor, 1-4-1 Nihonbashi,}, \city{Chuo-ku}, \postcode{1030027}, \state{Tokyo}, \country{Japan}}}


\abstract{This paper introduces quantum circuit $C^*$-algebra net, which provides a connection between $C^*$-algebra nets proposed in classical machine learning and quantum circuits.
Using $C^*$-algebra, a generalization of the space of complex numbers, we can represent quantum gates as weight parameters of a neural network.
By introducing additional parameters, we can induce interaction among multiple circuits constructed by quantum gates.
This interaction enables the circuits to share information among them, which contributes to improved generalization performance in machine learning tasks.
As an application, we propose to use the quantum circuit $C^*$-algebra net to encode classical data into quantum states, which enables us to integrate classical data into quantum algorithms.
Numerical results demonstrate that the interaction among circuits improves performance significantly in image classification, and encoded data by the quantum circuit $C^*$-algebra net are useful for downstream quantum machine learning tasks.
}




\maketitle

\section{Introduction}\label{sec:introduction}

$C^*$-algebra has been actively researched in quantum mechanics, mathematical physics, and pure mathematics~\cite{murphy90,lance95,bru23}.
It is a natural generalization of the space of complex numbers.
Typical examples include the space of operators and the space of continuous functions.
$C^*$-algebra was first proposed in quantum mechanics to model physical observables since they are represented by operators.
It has also been studied in pure mathematics.

Recently, applying $C^*$-algebras to machine learning methods has been investigated~\cite{hashimoto21,hashimoto23-aistats,hashimoto23-deeprkhm,hashimoto_postion}.
Using $C^*$-algebras, we can treat various types of data, such as functional data, image data, and graph data, in a unified way, since they can be described by functions and operators.
Although the application was originally focused on kernel methods, $C^*$-algebras are also introduced into the parameters of neural networks~\cite{hashimoto22}.
By focusing on the $C^*$-algebra of continuous functions and generalizing the neural network parameters to the $C^*$-algebra-valued ones, we can combine multiple models with real- or complex-valued parameters continuously and train them simultaneously~\cite{hashimoto22}.
Thus, a neural network parameterized such $C^*$-algebra values, known as (commutative) $C^*$-algebra net, can be regarded as multiple real- or complex-valued neural networks with interaction among them, and is applicable, for example, to ensemble learning~\cite{dong20,ganaie22} and few-shot learning~\cite{fei-fei06,lake11}.
However, because the product structure of the $C^*$-algebra of continuous functions is commutative, a special loss function is necessary to induce interactions among models during the training.
To resolve this limitation, commutative $C^*$-algebra net is further generalized to noncommutative $C^*$-algebra net, consisting of noncommutative parameters, such as non-diagonal matrices~\cite{hataya23_cstar}.
By virtue of the noncommutative product structure of the $C^*$-algebra, the multiple models can interact automatically and share information to obtain higher performance.

In the realm of quantum machine learning, quantum gates, represented by unitary matrices, can also be regarded as elements of the $C^*$-algebra of matrices.
This generalized viewpoint enables us to naturally handle quantum circuits using $C^*$-algebras and connect quantum circuits with $C^*$-algebra net.

In this paper, we propose quantum circuit $C^*$-algebra net, which is a $C^*$-algebra net motivated by quantum circuits.
We consider the $C^*$-algebra net over the $C^*$-algebra of matrices as quantum circuits by limiting its $C^*$-algebra-valued (i.e., matrix-valued) parameters of the network to unitary matrices to represent quantum gates.
This formulation provides a connection between quantum circuits and $C^*$-algebra nets.
We first focus on the case where there are multiple individual quantum circuits and they are separated.
We show that this setting corresponds to the commutative $C^*$-algebra net.
Then, we generalize the setting to the case where the circuits interact to share their information, which corresponds to the noncommutative $C^*$-algebra net.

As an application of our proposed quantum circuit $C^*$-algebra net, we consider encoding of classical data into quantum states.
We first transform classical data into quantum states using a standard neural network.
The unitary-matrix-valued parameters of the quantum circuit $C^*$-algebra net transform quantum states into better states to describe the data.
Using the quantum states obtained by the quantum circuit $C^*$-algebra net, we can input the data into quantum computers to apply quantum machine learning algorithms.
We also investigate an approach to encoding time-series data using the quantum circuit $C^*$-algebra net.

The validity of quantum circuit $C^*$-algebra net is demonstrated using real-world image datasets, such as MNIST~\cite{lecun2010mnist}.
Specifically, our experiments reveal that quantum circuit $C^*$-algebra nets with interaction outperform those without interaction in image classification tasks, indicating the power of noncommutative product structure.
Additionally, we show that the quantum states encoded by the quantum circuit $C^*$-algebra net are applicable to quantum machine learning with quantum kernels as the existing quantum encoding method, such as AQCE~\cite{shirakawa2021automatic}.

The remaining manuscript is structured as follows.
\Cref{sec:preliminaries} briefly introduces $C^*$-algebra and its application to neural networks, $C^*$-algebra net.
Then, the proposed quantum circuit $C^*$-algebra net is presented in \cref{sec:quantum_cstar_net}.
In \cref{sec:applications}, we discuss its applications to encoding of classical data into quantum states.
\Cref{sec:experiments} demonstrates that the quantum circuit $C^*$-algebra net with interaction improves the performance and the encoded states by the quantum circuit $C^*$-algebra net can be used in quantum machine learning.
Finally, \cref{sec:conclusion} concludes the manuscript with discussions on potential future directions.

\section{Preliminaries}\label{sec:preliminaries}
We review theoretical notions and existing works related to this paper.

\subsection{$C^*$-algebra}
$C^*$-algebra is a generalization of the space of complex values.
It has structures of the product, involution $^*$, and norm.
\begin{definition}[$C^*$-algebra]~\label{def:c*_algebra}
A set $\mcl{A}$ is called a {\em $C^*$-algebra} if it satisfies the following conditions:

\begin{enumerate}[itemsep=0pt,topsep=0pt]
 \item $\mcl{A}$ is an algebra over $\mathbb{C}$ and {equipped with} a bijection $(\cdot)^*:\mcl{A}\to\mcl{A}$ that satisfies the following conditions for $\alpha,\beta\in\mathbb{C}$ and $c,d\in\mcl{A}$:

\begin{itemize}[itemsep=0pt,topsep=0pt]
    \item $(\alpha c+\beta d)^*=\overline{\alpha}c^*+\overline{\beta}d^*$,
    \item $(cd)^*=d^*c^*$,
    \item $(c^*)^*=c$.
\end{itemize}

 \item $\mcl{A}$ is a normed space with $\Vert\cdot\Vert$, and for $c,d\in\mcl{A}$, $\Vert cd\Vert\le\Vert c\Vert\,\Vert d\Vert$ holds.
 In addition, $\mcl{A}$ is complete with respect to $\Vert\cdot\Vert$.

 \item For $c\in\mcl{A}$, $\Vert c^*c\Vert=\Vert c\Vert^2$ ($C^*$-property) holds.
\end{enumerate}
\end{definition}

The product structure in $C^*$-algebras can be both commutative and noncommutative.
\begin{example}[Commutative $C^*$-algebra]
Let $\alg$ be the space of $m$ by $m$ complex-valued diagonal squared matrices.
We can regard $\alg$ as a $C^*$-algebra by setting
\begin{itemize}[itemsep=0pt,topsep=0pt]
    \item Product: The product of two matrices, i.e., for $a_1,a_2\in\alg$, $(a_1a_2)_{i,i}=(a_1)_{i,i}(a_2)_{i,i}$ for $i=1,\ldots,m$ and $(a_1a_2)_{i,j}=0$ for $i,j=1,\ldots,m$ with $i\neq j$.
    Here, for a matrix $a$, $a_{i,j}$ is the $(i,j)$-entry of $a$.
    \item Involution: Hermitian conjugate, i.e., for $a\in\alg$, $(a^*)_{i,i}=\overline{a_{i,i}}$ for $i=1,\ldots,m$ and $(a^*)_{i,j}=0$ for $i,j=1,\ldots,m$ with $i\neq j$.
    \item Norm: Operator norm, i.e., for $a\in\alg$, $\Vert a\Vert=\max_{i=1,\ldots,m}\vert a_{i,i}\vert$.
\end{itemize}
In this case, the product in $\alg$ is commutative, that is, $a_1a_2=a_2a_1$ for $a_1,a_2\in\alg$.
\end{example}
\begin{example}[Noncommutative $C^*$-algebra]
Let $\alg$ be the space of $m$ by $m$ (not necessarily diagonal) complex-valued squared matrices.
We can regard $\alg$ as a $C^*$-algebra by setting
\begin{itemize}[itemsep=0pt,topsep=0pt]
    \item Product: The product of two matrices, i.e., for $a_1,a_2\in\alg$, $(a_1a_2)_{i,j}=\sum_{k=1}^m(a_1)_{i,k}(a_2)_{k,j}$ for $i,j=1,\ldots,m$.
    \item Involution: Hermitian conjugate, i.e., for $a\in\alg$, $(a^*)_{i,j}=\overline{a_{j,i}}$ for $i,j=1,\ldots,m$.
    \item Norm: Operator norm, i.e., for $a\in\alg$, $\Vert a\Vert=\max_{\Vert \ket{q}\Vert=1}\Vert a\ket{q}\Vert$.
    Here, for $\ket{q}\in\mathbb{C}^{m}$, $\Vert \ket{q}\Vert=\sqrt{\qbracket{q}{q}}$.
\end{itemize}
Here, $\Vert\cdot\Vert_{\hil}$ is the norm in $\hil$.
In this case, the product in $\alg$ is noncommutative.
Note that if $\hil$ is a $d$-dimensional space for a finite natural number $d$, then elements in $\alg$ are $d$ by $d$ matrices.
\end{example}

\subsection{$C^*$-algebra net}\label{subsec:c_algebra_nets}
\citet{hashimoto22} proposed generalizing real-valued neural network parameters to commutative $C^*$-algebra-valued ones.
Let $\alg=C(\mcl{Z})$, the commutative $C^*$-algebra of continuous functions on a compact Hausdorff space $\mcl{Z}$.
Let $L$ be the depth of the network and $d_0,\ldots,{d_{L}}$ be the width of each layer.
For $j=1,\ldots,L$, set $W_j:\alg^{d_{i-1}}\to\alg^{d_{i}}$ as a $d_{j}\times d_{j-1}$ $\alg$-valued matrix that corresponds to weight parameters.
In addition, set a nonlinear activation function $\sigma_j:\alg^{d_{j}}\to\alg^{d_{j}}$.
The commutative $C^*$-algebra net $f:\alg^{d_0}\to\alg^{d_{L}}$ is defined as
\begin{align*}
 f(x)=\sigma_{L}(W_L\cdots\sigma_1(W_1x)\cdots).
\end{align*}
By generalizing neural network parameters to functions, we can combine multiple standard (real-valued) neural networks continuously, which enables us to train them simultaneously and efficiently.

If $\mcl{Z}$ is a finite set with $m$ elements, then we have $\alg=\{a\in \mathbb{C}^{m\times m}\,\mid\, a\mbox{ is a diagonal matrix}\}$.
The single $C^*$-algebra net $f$ on $\alg$ corresponds to $m$ separate real or complex-valued sub-models.
Indeed, denote by $x^J$ the vector composed of the $J$th diagonal elements of $x\in\alg^N$, which is defined as the vector in $\mathbb{C}^{N}$ whose $k$th element is the $J$th diagonal element of the $k$th element (note that $k$th element of $x$ is a matrix) of $x$.
Assume the activation function $\sigma_j:\alg^N\to\alg^N$ is defined as $\sigma_j(x)^J=\tilde{\sigma}_j(x^J)$ for some $\tilde{\sigma}_j:\mathbb{C}^N\to\mathbb{C}^N$.
Since the $J$th diagonal element of $a_1a_2$ for $a_1,a_2\in\alg$ is the product of the $J$th element of $a_1$ and $a_2$, we have
\begin{align}
f(x)^J=\tilde{\sigma}_{L}(W_{L}^J\cdots\tilde{\sigma}_1(W_{1}^Jx)\cdots),\label{eq:c_net_diagonal}
\end{align}
where $W_{j}^J\in\mathbb{C}^{d_j\times d_{j-1}}$ is the matrix whose $(k,l)$-entry is the $J$th diagonal of the $(k,l)$-entry of $W_j\in\alg^{d_j\times d_{j-1}}$.

Hataya and Hashimoto~\cite{hataya23_cstar} proposed to generalize the $C^*$-algebra from $\{a\in \mathbb{C}^{m\times m}\,\mid\, a\mbox{ is a diagonal matrix}\}$ to $\mathbb{C}^{m\times m}$, the space of (not necessarily diagonal) $m$ by $m$ matrices.
In this case, we cannot separate the $C^*$-algebra net to $m$ sub-models like Eq.~\eqref{eq:c_net_diagonal}.
However, we can still interpret that we have $m$ sub-models, but the non-diagonal elements of the matrix-valued parameters induce the interaction among the sub-models.
Using this interaction, sub-models cooperate to improve performance: each sub-model can access and learn from information from other sub-models, which is otherwise impossible to use.
Its potential application lies in peer-to-peer machine learning~\cite{vanhaesebrouck2017decentralized,bellet2018personalized}, where each sub-model is located separately and communicates with others automatically.

\section{Quantum circuit $C^*$-algebra net}\label{sec:quantum_cstar_net}
We propose a new type of $C^*$-algebra net called quantum circuit $C^*$-algebra net, which is motivated by quantum circuits.
We first propose a simple $C^*$-algebra net without interactions among circuits; then we generalize it to induce interactions.

\subsection{Quantum circuit $C^*$-algebra net without interactions}\label{subsec:qc_net_nointeraction}
We consider an $L$-layer $C^*$-algebra net that is motivated by quantum circuits.
We consider an $m$-state system (typically, $m=2^l$ for some $l\in\mathbb{N}$), and let $\alg=\mathbb{C}^{m\times m}$.
Let $d$ be the number of circuits (corresponding to the width of the network), and for $j=1,\ldots,L$, 
let $u^j_1,\ldots,u^j_{d}\in\alg$ be unitary matrices corresponding to quantum gates.
Let $\ket{q_1},\ldots,\ket{q_d}\in\mathbb{C}^{m}$ be $d$ initial states encoding data.
In addition, let $C\in\mathbb{N}$ be the number of measurements for each circuit, and for $i=1,\ldots,C$, let $\tilde{\eta}_i:\mathbb{C}^m\to\mathbb{R}$ be a nonlinear map corresponding to the measurement.
For example, $\eta_i$ is defined as $\eta_i(\ket{q_k})=\bra{q_k}a_i\ket{q_k}$ for some Hermitian matrix $a_i\in\alg$ corresponding to the observable, and $a_i$ can be regarded as learnable parameters.
We set $\tilde{\eta}:\mathbb{C}^m\to\mathbb{R}^C$ as $\tilde{\eta}(\ket{q_k})=(\tilde{\eta}_1(\ket{q_k}),\ldots,\tilde{\eta}_C(\ket{q_k}))$.
Consider a constant-width $C^*$-algebra net
\begin{align}
f^{\theta}(\ket{q})=\eta(U^L\cdots U^1\ket{q}),\label{eq:qc_net_nointeraction}
\end{align}
where for $j=1,\ldots,L$, $U^j\in\alg^{d\times d}$ is the $d$ by $d$ $\alg$-valued matrix whose diagonal elements are $u^j_1,\ldots,u^j_{d}$ and nondiagonal elements are all $0$.
In addition, $\ket{q}\in (\mathbb{C}^m)^{d}$ is the $d$-dimensional $\mathbb{C}^m$-valued vector whose elements are $\ket{q_1},\ldots,\ket{q_{d}
}$, 
$\eta:(\mathbb{C}^m)^{d}\to (\mathbb{R}^{C})^d$ is defined as $\eta(\ket{q})=(\tilde{\eta}(\ket{q_1}),\ldots,\tilde{\eta}_{C}(\ket{q_d}))$, and $\theta=(U^1,\ldots,U^L)$ is the aggregation of learnable parameters.
Then, the network $f^{\theta}$ corresponds to $d$ individual quantum circuits represented as
\begin{align*}
\tilde{\eta}(u^1_k\cdots u^L_k\ket{q_k})\quad (k=1,\ldots,d).
\end{align*}
We call the network~\eqref{eq:qc_net_nointeraction} the \emph{quantum circuit $C^*$-algebra net}.

Although this quantum circuit $C^*$-algebra net consists of $d$ circuits, they are independent and do not interact with each other.
In other words, when a state $\ket{q_k}$ is given to the $k$th circuit for $k=\{1,\dots, d\}$, its information does not provide any effects to other circuits.


\subsection{Quantum circuit $C^*$-algebra net with interactions}
In the network~\eqref{eq:qc_net_nointeraction}, each quantum circuit $\tilde{\eta}(u^L_k\cdots u^1_k\ket{q_k})$ is constructed individually and does not interact with other circuits.
To induce the interactions among circuits, we generalize the network~\eqref{eq:qc_net_nointeraction} as follows:
For $j=1,\ldots,L$, let $d_j\in\mathbb{N}$ be the width of the $j$th layer.
For $i=1,\ldots,d_{j}$ and $k=1,\ldots,d_{j-1}$, let $u^j_{i,k}\in\alg$ be a unitary matrix for $i=k$ and a general matrix for $i\neq k$.
Then, we define $U^j\in\alg^{d_{j-1}\times d_j}$ as an $\alg$-valued matrix whose $(i,k)$-element is $u^j_{i,k}\in\alg$.
In addition, for $j=1,\ldots,L$, let $\sigma_j:\alg^{d_j}\to\alg^{d_j}$ be a nonlinear map.
For example, $\sigma_j(\ket{q})=(\ket{q_1}/\Vert \ket{q_1}\Vert,\ldots,\ket{q_{d_j}}/\Vert \ket{q_{d_j}}\Vert)$ for $\ket{q}=(\ket{q_1},\ldots,\ket{q_{d_j}})$ since $\ket{q_i}$ is no longer normalized.
We define $\eta:(\mathbb{C}^m)^{d_L}\to(\mathbb{R}^{C})^{d_L}$ in the same manner as that in Subsection~\ref{subsec:qc_net_nointeraction}.
Then, we obtain a $C^*$-algebra net
\begin{align}
f^{\theta}(\ket{q})=\eta(\sigma_L(U^L\cdots \sigma_1(U^1\ket{q})\cdots)).\label{eq:qc_net_interaction}
\end{align} 
The nondiagonal elements in $U^j$ induce interactions with other circuits.
Specifically, when an initial state $\ket{q_k}$ is provided to the $k$th circuit for $k=\{1,\dots, d\}$, other circuits can also learn from its information.
Suppose we have $d$ mutually inclusive sets of quantum states and train $d$ circuits independently with different subsets.
Even though each circuit can only access the corresponding subset, it can indirectly receive information from other circuits due to interaction, yielding better generalization performance.

\subsection{Connection with commutative and noncommutative $C^*$-algebra nets}\label{subsec:noncommutative}
As we discussed in Subsection~\ref{subsec:c_algebra_nets}, commutative and noncommutative $C^*$-algebra nets are proposed.
Although the noncommutative $C^*$-algebra nets can induce interactions among sub-models, the commutative $C^*$-nets do not. 
The quantum circuit $C^*$-algebra nets with and without interactions are also interpreted as noncommutative and commutative $C^*$-algebra nets.
Let $P$ be the permutation matrix that satisfies $P\ket{e_{im+j}}=\ket{e_{j{d}+i}}$, where $\ket{e_{k}}$ is the unit vector in $\mathbb{C}^{md}$ whose $k$th element is $1$ and all the other elements are $0$.
Let $U^j$ be the $\alg^{d\times d}$-valued matrix that is defined in Subsection~\ref{subsec:qc_net_nointeraction}.
Let $\ket{q'}=U^j\ket{q}$.
Then, we have
\begin{align*}
(PU^jP^*)P\ket{q}&=P\ket{q'}.
\end{align*}
By the definition of $P$, $PUP^*$ is an $m$ by $m$ matrix each of whose element is a $d$ by $d$ diagonal matrix.
Therefore, the network~\eqref{eq:qc_net_nointeraction} is regarded as a commutative $C^*$-algebra net over the $C^*$-algebra of squared diagonal matrices.
On the other hand, if $d_j=d$ for $j=1,\ldots,L$, then the matrix $U^j$ in the network~\eqref{eq:qc_net_interaction} is transformed into an $m$ by $m$ matrix each of whose element is a $d$ by $d$ full matrix by $P$.
Therefore, the network~\eqref{eq:qc_net_interaction} is regarded as a noncommutative $C^*$-algebra net over the $C^*$-algebra of general squared matrices.
This correspondence is the same as the existing study of commutative and noncommutative $C^*$-algebra nets by means of interactions.

\section{Applications}\label{sec:applications}
We provide several applications of the proposed quantum circuit $C^*$-algebra net to encode classical data into quantum states.
\subsection{Encoding data to quantum states}\label{subsec:encoder}
We can combine the quantum circuit $C^*$-algebra with a standard neural network to obtain quantum states that describe data.
We set $d_0=d_L$ here.
Let $e_0\in\mathbb{N}$ be the input dimension, and for $k=1,\ldots,d$, let $x_k\in\mathbb{R}^{e_0}$ be the input vector for the $k$th circuit.
Let $g^{\xi}:\mathbb{R}^{e_0}\to \mathbb{C}^m$ be a (standard) neural network that maps the input vector $x_k\in\mathbb{R}^{e_0}$ to a corresponding quantum state $\ket{q_k}\in\mathbb{C}^m$, where $\xi$ is the vector that aggregates the learnable parameters of the neural network.
We set $\ket{q}=(\ket{q_1},\ldots,\ket{q_{d}})$.
Then, we transform the state $\ket{q}$ using a quantum circuit $C^*$-algebra net $f^{\theta}$ with $C=e_0$.
Using $f^{\theta}$, we obtain another quantum state $\ket{q'}=\sigma_L(U^L\cdots \sigma_1(U^1\ket{q})\cdots) $ and finally get a vector in $(\mathbb{R}^{e_0})^{d}$ by the final nonlinear map $\eta$.
We learn $g^{\xi}$ and $f^{\theta}$ so that for $k=1,\ldots, d_L$, $x_k$ and the output $\big(f^{\theta}(g^{\xi}(x_1),\ldots,g^{\xi}(x_{d}))\big)_k$ of the $k$th circuit are close.

\begin{figure}
    \centering
    \includegraphics[scale=0.4]{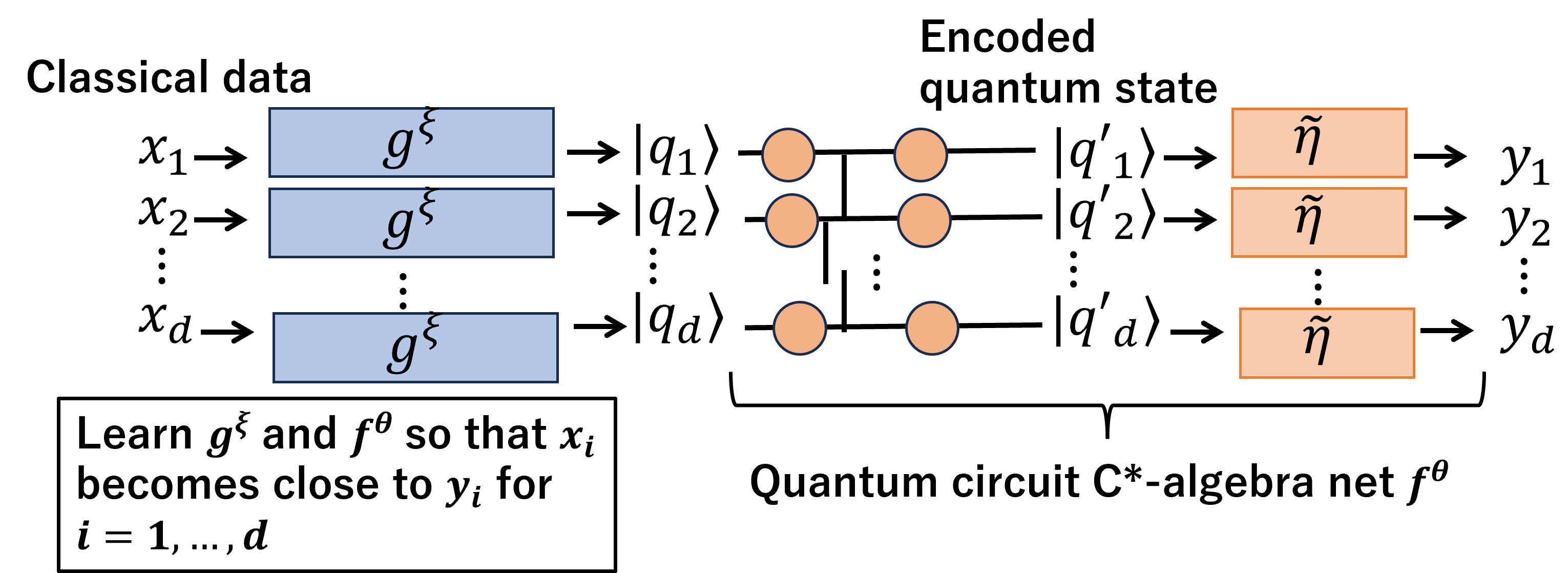}
    \caption{Overview of quantum circuit $C^*$-algebra net for encoding classical data to quantum states}
    \label{fig:encoding}
\end{figure}

The network $\tilde{f}^{\theta}:(\mathbb{R}^{e_0})^{d}\to(\mathbb{C}^m)^{d}$ defined as 
\begin{align*}
   \tilde{f}^{\theta}(x)=\sigma_L(U^L\cdots \sigma_1(U^1(g^{\xi}(x_1),\ldots,g^{\xi}(x_d))\cdots) 
\end{align*}
can be regarded as an encoder that encodes the input vector $x$ to a quantum state $\ket{q'}$.
If we want to input data into a quantum computer, we have to transform the data into quantum states.
For the classical input data $x_k$, we can input $\ket{q'_k}$ to the quantum computer.
Figure~\ref{fig:encoding} schematically shows how the quantum $C^*$-algebra net can be used for encoding classical data to quantum states.

\subsection{Encoding time-series data to a quantum state series}

\begin{figure}
    \centering
    \includegraphics[scale=0.45]{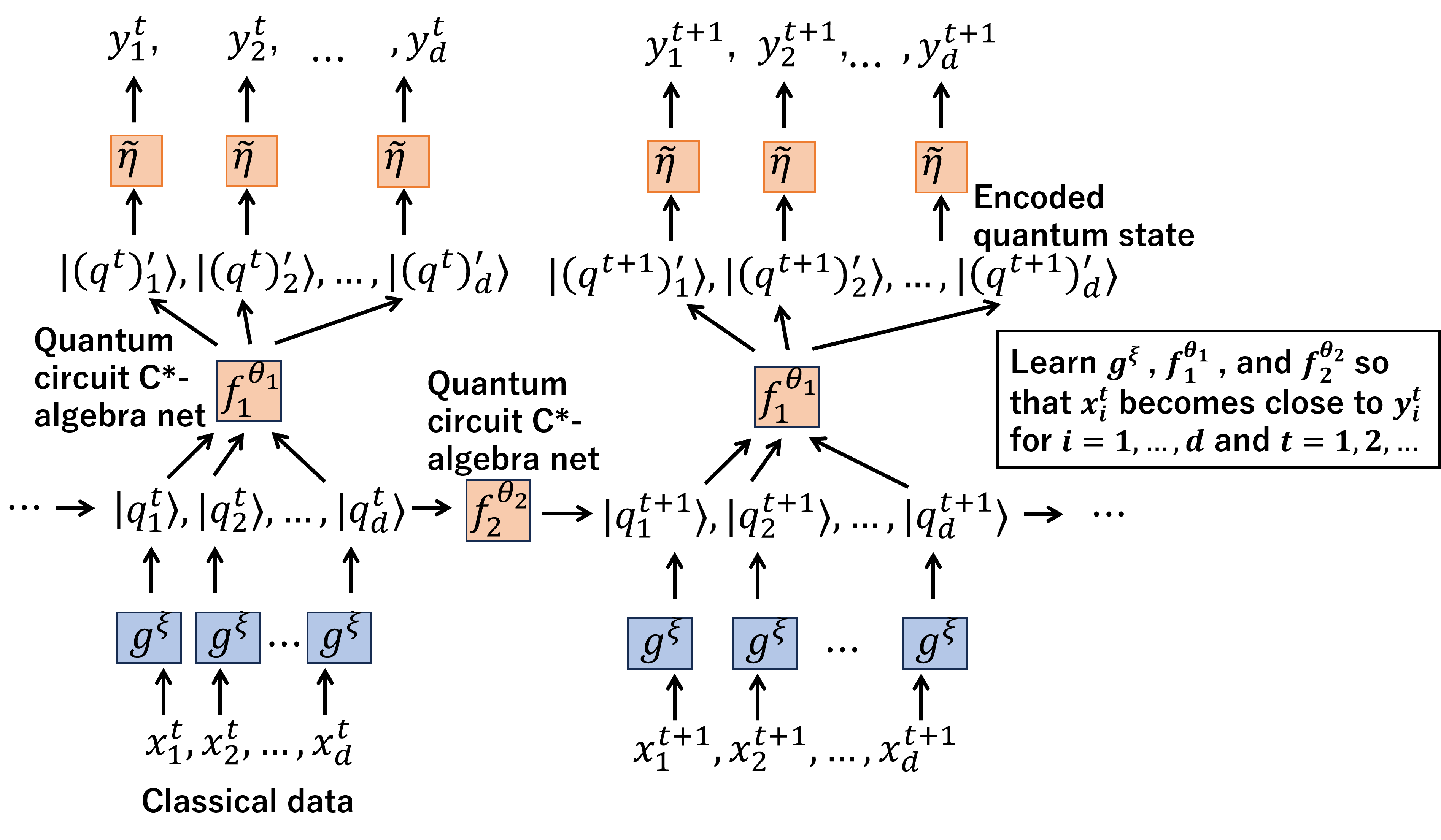}
    \caption{Overview of quantum circuit $C^*$-algebra net for encoding classical time-series data to quantum states}
    \label{fig:time-series}
\end{figure}

We can also apply the encoding method proposed in Subsection~\ref{subsec:encoder} to recurrent neural networks (RNN)~\cite{elman90} for time-series data.
Let $e_0\in\mathbb{N}$ be the input dimension, and for $k=1,\ldots,d$, let $x_k^1,x_k^2,\ldots\in\mathbb{R}^{e_0}$ be the input time-series.
Analogous to the Subsection~\ref{subsec:encoder}, let $g^{\xi}:\mathbb{R}^{e_0}\to \mathbb{C}^m$ be a (standard) neural network that maps the input vector $x_k^t$ to a corresponding quantum state $\ket{\hat{q}_k^t}$.
We consider a quantum circuit $C^*$-algebra net $f_1^{\theta_1}:(\mathbb{C}^{m})^{d}\to(\mathbb{R}^C)^{d}$ and a quantum circuit $C^*$-algebra net $\tilde{f}_2^{\theta_2}:(\mathbb{C}^{m})^{d}\to (\mathbb{C}^{m})^{d}$ without the final nonlinear map $\eta$.
The quantum circuit $C^*$-algebra net
$\tilde{f}_2^{\theta_2}$ maps the quantum state $\ket{q_k^t}$ at time $t$ to the quantum state $\ket{\tilde{q}_k^{t+1}}$ at time $t+1$.
Then, we aggregate the outputs of $g^{\xi}$ and $f_2^{\theta_2}$ and obtain $\ket{q^{t+1}_k}$ as $\ket{q^{t+1}_k}=(\ket{\tilde{q}_k^{t+1}}+\ket{\hat{q}_k^{t+1}})/\sqrt{2}$.
On the other hand, $f_1^{\theta_1}$ takes the same role as $f^{\theta}$ in Subsection~\ref{subsec:encoder}.
We transform $\ket{q^t}=(\ket{q^t_1},\ldots,\ket{q^t_d})\in(\mathbb{C}^m)^d$ into $\ket{(q^t)'}$ by $\ket{(q^t)'}=\sigma_{1,L}(U^{1,L}\cdots \sigma_{1,1}(U^{1,1}\ket{q^t})\cdots)$ and learn $g^{\xi}$, $f_1^{\theta}$, and $f_2^{\theta}$ so that $x_k^t$ and the $k$ the element of the output $f_1^{\theta_1}(\ket{q^t})\in(\mathbb{R}^C)^d$ are close for $t=1,2\ldots$ and $k=1,\ldots,d$.
Here, we denote the unitary matrix in $f_1^{\theta_1}$ by $U^{1,j}$ and the nonlinear map in $f_1^{\theta_1}$ by $\sigma_{1,j}$.
In summary, the quantum circuit $C^*$-algebra RNN is
\begin{align*}
\ket{\hat{q}^t}&=(g^{\xi}(x^t_0),\ldots g^{\xi}(x^t_d))\\
\ket{\tilde{q}^{t}}&=f_2^{\theta_2}(\ket{q^{t-1}})\\
\ket{q^{t}}&=\frac{1}{\sqrt{2}}(\ket{\tilde{q}^{t}}+\ket{\hat{q}^{t}})\\
\ket{(q^t)'}&=\sigma_{1,L}(U^{1,L}\cdots \sigma_{1,1}(U^{1,1}\ket{q^t})\cdots)\\
y^t&=\eta_1(\ket{(q^t)'})\quad (y^{t}=f_1^{\theta_1}(\ket{q^t})).
\end{align*}
The quantum state $\ket{(q^t)'_k}$ describes the data $x^t_k$ at time $t$.
Figure~\ref{fig:time-series} schematically shows how the quantum $C^*$-algebra net can be used for encoding classical data to quantum states.

\section{Experiments}\label{sec:experiments}

We present the empirical validity of the proposed quantum circuit $C^*$-algebra net with real-world datasets.

\begin{table}[t]
    \centering
    \begin{tabular}{lccccc}
    \toprule
            & w/o interaction & \multicolumn{4}{c}{w/ interaction} \\
     Dataset       &        &  4 circuits  & 8 circuits   &  16 circuits   & 32 circuits   \\
    \cmidrule(lr){1-1} \cmidrule(lr){2-2} \cmidrule(lr){3-6}
    MNIST            & 0.8354       &  0.9395                   & 0.9512     & 0.9628      & 0.9620     \\
    Fashion-MNIST    & 0.3962       &  0.8359                   & 0.8480     & 0.8594      & 0.8600     \\
    Kuzushiji-MNIST  & 0.3572       &  0.7308                   & 0.7883     & 0.8110      & 0.8128     \\
    \bottomrule
    \end{tabular}
    \caption{Test accuracy of quantum circuit $C^*$-algebra nets with and without interaction. The number of circuits corresponds to $d$.}
    \label{tab:comparison}
\end{table}

\begin{table}[t]
    \centering
    \begin{tabular}{lccc}
    \toprule
                      &  AQCE     &  \multicolumn{2}{c}{Quantum circuit $C^*$-algebra net}  \\
    Dataset           &           &   Normalization              & Scaling   \\
    \cmidrule(lr){1-1} \cmidrule(lr){2-2} \cmidrule(lr){3-4}
    MNIST          & 0.9486    &    0.9190     &   0.9638           \\
    Fashion-MNIST  & 0.8204    &    0.8070     &   0.8297         \\
    Kuzushiji-MNIST& 0.8417    &    0.7683     &   0.8356       \\
    \bottomrule
    \end{tabular}
    \caption{Test accuracy of SVM with quantum kernels constructed with states encoded by AQCE and quantum circuit $C^*$-algebra net. For quantum circuit $C^*$-algebra nets, the results with activation functions of normalization and scaling are presented.}
    \label{tab:encoded_classification}
\end{table}

\subsection{Quantum circuit $C^*$-algebra net}\label{subsec:exp_quantum_cstar_net}

First, we present the empirical results of quantum circuit $C^*$-algebra nets with and without interaction introduced in \cref{sec:quantum_cstar_net}.
Image datasets of MNIST~\cite{lecun2010mnist}, Fashion MNIST~\cite{xiao2017fashion}, and Kuzushiji MNIST~\cite{clanuwat2018deep} are adopted, which consist of pairs of images of $28\times 28$ pixels and $10$-category labels.
Each image $x$ is encoded into vectors $\ket{q}\in\mathbb{C}^{m}$ by using a multi-layer perception (MLP), where $m$ is set to $128$ so that each circuit can be represented by $l=7$ qubits.
The MLP is shared among all the circuits.
The quantum circuit $C^*$-algebra net is equipped with parameters in the space of unitary matrices and a normalizing activation function $\sigma_j(\ket{q})=(\ket{q_1}/\Vert\ket{q_1}\Vert,\ldots,\ket{q_{d}}/\Vert\ket{q_{d}}\Vert)$ presented in \cref{sec:quantum_cstar_net}.
Because such a space is the Stiefel manifold $\{U\in\mathbb{C}^{m\times m}\mid U^*U=I\}$~\cite{boumal2023intromanifolds}, we optimize the parameters using the Adam optimizer~\cite{Kingma2015} on the manifold, as explained in \cref{ap:sec:stiefel_optimization}.

The observation function $\eta$ for a quantum state $\ket{q'}$ consists of learnable projection matrices $\ket{p_i} \bra{p_i}$ for $i=1,\dots, C$, where $C$ is the number of categories (that is, $10$), and $\ket{p_i}$ is orthogonal so that $\qbracket{p_i}{p_j}=\delta_{ij}$.
The matrix of $P=[\ket{p_1}, \dots, \ket{p_C}]$ is also optimized with the Adam optimizer on the Stiefel manifold.
This observation function is also shared among circuits.
Each circuit is trained independently with different subsets of datasets: given a one-hot label $z\in\{0,1\}^C$ to an input data $x$, parameters are updated to minimize the mean squared loss $\sum_{i=1}^C\|\bra{q'_k}(|p_i\rangle\langle p_i|)\ket{q'_k}-z_i\|^2$ when the input data is provided for the $k$th circuit.
Further details on the experimental setup are presented in \cref{ap:sec:exp_details}.

\Cref{tab:comparison} shows the test accuracy of quantum circuit $C^*$-algebra nets with and without interaction, which averages per-circuit test accuracy.
For quantum circuit $C^*$-algebra nets with interactions, we present the results with $d=4, 8, 16, 32$ circuits.
The quantum circuit $C^*$-algebra net without interaction has the same number of learnable parameters with that with interaction consisting of 4 circuits.
Each circuit is optimized on mutually exclusive 1,000 data points.
From \cref{tab:comparison}, we can observe that the interaction induced by quantum circuit $C^*$-algebra net contributes to performance.
Firstly, under the fixed number of learnable parameters, i.e., w/o interaction and w/ interaction consisting of 4 circuits, the performance gain is significant.
Furthermore, test accuracy improves as the number of circuits increases, showing that each circuit can virtually access more data thanks to interaction.
These results demonstrate the effectiveness of interaction derived from noncommutativity of the product structure in the $C^*$-algebra described in Subsection~\ref{subsec:noncommutative}.

\subsection{Data encoding by quantum circuit $C^*$-algebra net}\label{subsec:exp_encoding}

Next, we show the numerical results of quantum kernels constructed with quantum states encoded by quantum circuit $C^*$-algebra net, explained in \cref{subsec:encoder}.
We use the same image datasets and a similar network with a single circuit ($d=1$) in \cref{subsec:exp_quantum_cstar_net}.
Exceptionally, $m=1024$ and $C=1024$, and parameters are updated to minimize the mean squared loss $\sum_{i=1}^{C}\|\bra{q'}(|p_i\rangle\langle p_i|)\ket{q'}-x_i\|^2$, where $x_i$ is the $i$th pixel of a vectorized input image $x$, zero-padded to be 1024 dimensional vectors.
Quantum kernels $k(q_i, q_j)=|\qbracket{q_i}{q_j}|^2$~\cite{havlivcek2019supervised,kusumoto2021experimental} is constructed using the entire training data and is adopted to classify test image data using a support vector machine with the obtained quantum kernels in a 1-vs-1 manner.
Further details on the experimental setup are presented in \cref{ap:sec:exp_details}.

\Cref{tab:encoded_classification} displays test accuracy.
As a baseline, we present the results with quantum kernels computed from states of the same datasets encoded by Automatic Quantum Circuit Encoding (AQCE)~\cite{placidi2023mnisq,shirakawa2021automatic}.
As can be observed, the results with quantum circuit $C^*$-algebra net are comparable with the ones with AQCE.
We additionally display the results obtained from the quantum circuit $C^*$-algebra net with the scaling activation function, i.e., $\ket{q}\mapsto \ket{q}/\sqrt{m}$, which are comparable or even superior to the baseline.
These results indicate that the quantum circuit $C^*$-algebra net can effectively encode necessary information into quantum states.
At the same time, the discrepancies based on the choice of activation functions imply its necessity.

\section{Conclusion and discussion}\label{sec:conclusion}

In this paper, we presented quantum circuit $C^*$-algebra net that connects $C^*$-algebra net with quantum circuits.
Specifically, quantum circuit $C^*$-algebra net with interaction is introduced as a special case of noncommutative $C^*$-algebra net of matrices.
The numerical experiments demonstrated its effectiveness that multiple circuits cooperate and share information to achieve better generalization performance.
Additionally, the application of quantum circuit $C^*$-algebra net in the encoding of classical data into quantum states is presented.
Experiments indicated that it could obtain high-quality quantum states for downstream quantum machine learning tasks.

In this work, we constructed quantum circuit $C^*$-algebra nets on a single classical computer.
An interesting future direction of quantum circuit $C^*$-algebra nets with interactions is decentralized peer-to-peer machine learning \cite{vanhaesebrouck2017decentralized,bellet2018personalized}.
In this setting, each local circuit learns without sharing data with others in remote locations, while improving its ability by leveraging other circuits' information through communication, i.e., interaction among circuits.
The quantum $C^*$-algebra net provides a way of communication based on the quantum circuits.

From a more technical point of view, studying the effects of nonlinear activation functions would also be essential.
For the experiments in Section 5, we set the activation function $\sigma_j$ as a normalizing function $\sigma_j(\ket{q})=(\ket{q_1}/\Vert\ket{q_1}\Vert,\ldots,\ket{q_{d_j}}/\Vert\ket{q_{d_j}}\Vert)$ or a scaling function $\sigma_j(\ket{q})=\ket{q}/\sqrt{m}$.
Adding nonlinear effects to quantum circuits is already proposed~\cite{holmes23}, and the nonlinear activation functions correspond to this type of nonlinear effect.
As seen in the experiments, the choice of activation functions is crucial, and exploring appropriate nonlinearity for the quantum circuit $C^*$-algebra net is remained as another future work.

\begin{appendices}

\section{Optimization on the Stiefel Manifold}\label{ap:sec:stiefel_optimization}

Optimization of neural network parameters on the Stiefel manifold has been investigated in the machine learning field for decades~\cite{nishimori2005learning,wisdom2016full,li2020efficient}.
The Stiefel manifold is defined as $\mathcal{S}_d=\{U\in\mathbb{C}^{d\times d}|U^*U=I\}$, where $I$ is the identity matrix.
For a matrix $U\in\mathcal{S}_d$, any matrix $Z$ in its tangent space $\mathcal{T}_U\mathcal{S}_d$ satisfies $Z^*U-U^*Z=0$.
With the inner product in the tangent space $\langle Z, Z'\rangle=\mathrm{tr}\{Z^*(I-\frac{1}{2}UU^*)Z'\}$ for $Z,Z'\in \mathcal{T}_U\mathcal{S}_d$, the gradient in this manifold for the loss function $\ell$ is $AU$, where $A=G^*U-U^*G$ and $G=\frac{\partial \ell}{\partial U}$.
Then, the gradient descent on the Stiefel manifold using the Cayley transformation~\cite{boumal2023intromanifolds,wisdom2016full,li2020efficient} can be written as
\begin{equation}
    U\gets(I+\frac{\lambda}{2}A)^{-1}(I-\frac{\lambda}{2}A)U\label{ap:eq:cayley_transformation},
\end{equation}
where $\lambda>0$ is the learning rate.
This can be extended to adaptive optimizers, such as Adam.
We opt for the Adam optimizer on the Stiefel manifold with the Cayley transformation proposed in \cite{li2020efficient} with a slight modification to use matrix inversion as \cref{ap:eq:cayley_transformation} rather than their approximation, because we only consider smaller matrices, up to 1024 by 1024.

\section{Experimental Details}\label{ap:sec:exp_details}

We implemented neural networks including quantum circuit $C^*$-algebra nets with PyTorch v2.2~\cite{paszke2019pytorch}.

For the experiments in \cref{subsec:exp_quantum_cstar_net}, each image is encoded into a 128-dimensional complex vector representing a corresponding quantum state by a multi-layer perceptron (MLP) with four layers with the leaky-ReLU activation function, consisting of real-valued parameters.
The width of each layer is set to 256, and the MLP creates a pair of 128-dimensional real vectors, which is treated as a 128-dimensional complex vector.
Its output $\ket{q}$ is fed into a four-layer quantum circuit $C^*$-algebra net, composed of complex-valued parameters.
Then, the resultant vector $\ket{q'}$ is converted with learnable projection matrices.
The parameters are updated for 50 epochs for quantum $C^*$-algebra nets with interaction and 100 epochs otherwise using the Adam optimizer with a minibatch size of 32 for each circuit.
The learning rate is set to $1.0$ for the parameters on the Stiefel manifold, otherwise $10^{-4}$.

Similarly, the input for the experiments in \cref{subsec:exp_encoding} is encoded to a 1024-dimensional complex vector by the MLP with a depth of four and a width of 512.
Except for these dimensions, we adopt the same setting as the description above.
The baseline states are obtained from the MNIST, Fashion-MNIST, and Kuzushiji-MNIST datasets in MNISQ with fidelity of 80~\cite{placidi2023mnisq}.
A support vector machine classifier is adopted from scikit-learn~\cite{pedregosa2011scikit} with a squared l2 penalty of $1.0$ as the experiments in \cite{placidi2023mnisq}.




\end{appendices}


\subsection*{Author contributions}
Y. H. and R. H. contributed equally. Y. H. established the theoretical aspect of the method and R. H. conducted the experiments.

\subsection*{Funding}
R. H. was supported by JSPS KAKENHI Grant Number 23K20015.

\bibliography{c_star_ref}

\end{document}